\documentclass[letterpaper, 10 pt, conference]{ieeeconf}  % Comment this line out if you need a4paper

\IEEEoverridecommandlockouts                              % This command is only needed if 
\usepackage{graphicx}
\usepackage{xcolor}
\usepackage[linesnumbered,ruled,vlined]{algorithm2e}
\usepackage{amsmath}
\usepackage{amssymb}
\usepackage{hyperref}
\hypersetup{
    colorlinks=true,
    linkcolor=blue,
    filecolor=magenta,      
    urlcolor=cyan,
    citecolor=blue,
}
\usepackage{float}
\usepackage{graphicx}
\usepackage{subcaption}
\usepackage{caption}
\usepackage{siunitx}
\usepackage{booktabs}
\usepackage{tabularx}
\let\labelindent\relax
\usepackage[shortlabels, inline]{enumitem}

\newcommand{\tetra}{\text{TetraGrip}\xspace}

\title{\LARGE \bf
\tetra: Sensor-Driven Multi-Suction Reactive Object Manipulation in Cluttered Scenes
}

\author{Paolo Torrado$^{1}$, Joshua Levin$^{1}$, Markus Grotz$^{2}$ and Joshua Smith$^{1,2}$
\thanks{$^{1}$ Electrical and Computer Engineering Department, University of Washington}%
\thanks{$^{2}$ Paul G. Allen School of Computer Science \& Engineering, University of Washington}%,{\tt\small b.d.researcher@ieee.org}
}

\begin{document}

\maketitle
\thispagestyle{empty}
\pagestyle{empty}

%%%%%%%%%%%%%%%%%%%%%%%%%%%%%%%%%%%%%%%%%%%%%%%%%%%%%%%%%%%%%%%%%%%%%%%%%%%%%%%%

\begin{abstract}

Warehouse robotic systems equipped with vacuum grippers must reliably grasp a diverse range of objects from densely packed shelves. However, these environments present significant challenges, including occlusions, diverse object orientations, stacked and obstructed items, and surfaces that are difficult to suction. We introduce \tetra, a novel vacuum-based grasping strategy featuring four suction cups mounted on linear actuators. Each actuator is equipped with an optical time-of-flight (ToF) proximity sensor, enabling reactive grasping.

We evaluate \tetra in a warehouse-style setting, demonstrating its ability to manipulate objects in stacked and obstructed configurations. Our results show that our RL-based policy improves picking success in stacked-object scenarios by 22.86\% compared to a single-suction gripper. Additionally, we demonstrate that TetraGrip can successfully grasp objects in scenarios where a single-suction gripper fails due to physical limitations, specifically in two cases: (1) picking an object occluded by another object and (2) retrieving an object in a complex scenario. These findings highlight the advantages of multi-actuated, suction-based grasping in unstructured warehouse environments. The project website is available at: \href{https://tetragrip.github.io/}{https://tetragrip.github.io/}.

\end{abstract}

%%%%%%%%%%%%%%%%%%%%%%%%%%%%%%%%%%%%%%%%%%%%%%%%%%%%%%%%%%%%%%%%%%%%%%%%%%%%%%%%
\section{Introduction}

\begin{figure}[H]
    \raisebox{-20mm}{\includegraphics[trim=0.cm 2.cm 5.cm 0.cm, clip, width=.59\textwidth]{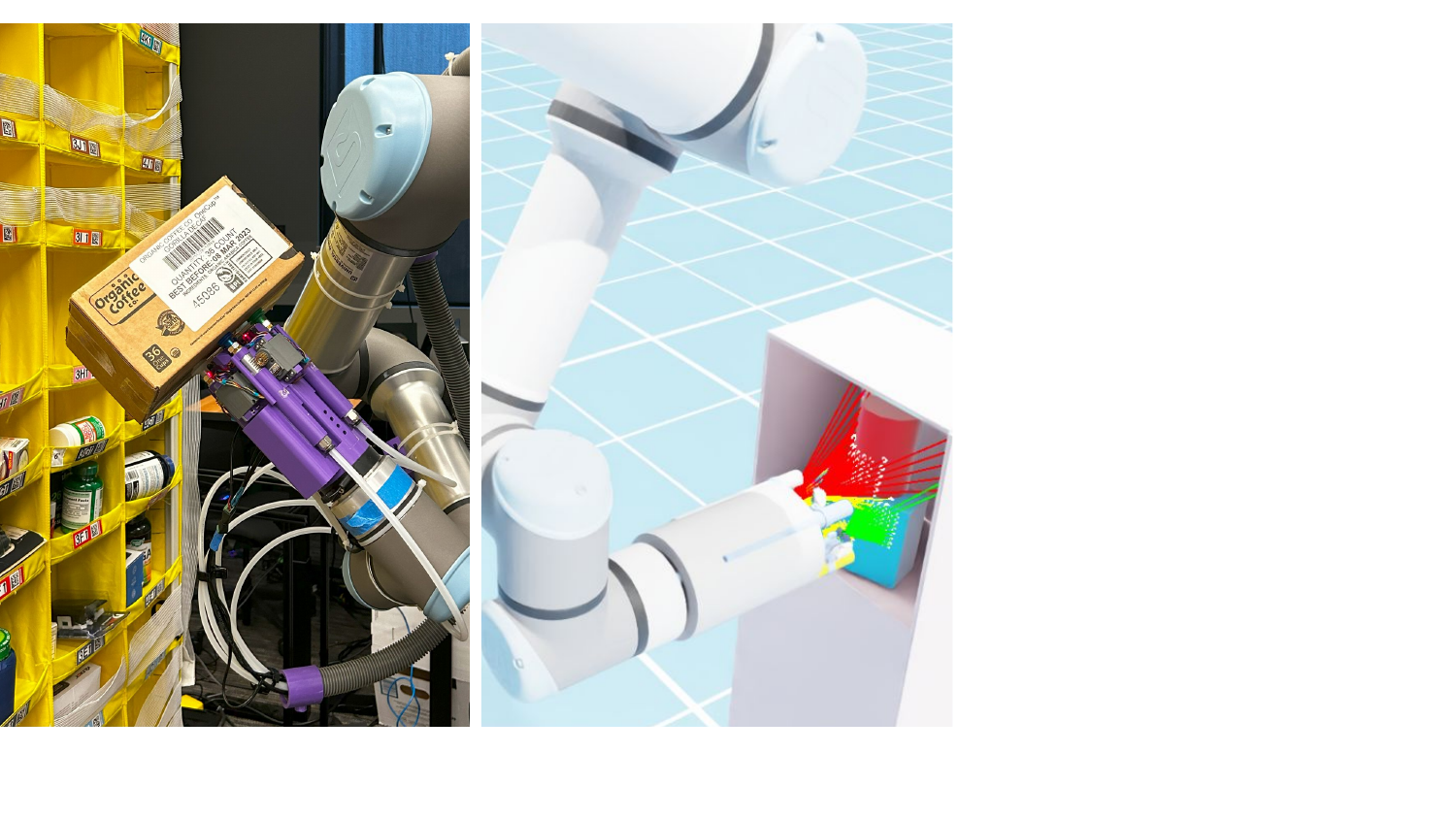}}
    \caption{(Left) \tetra picking an item from an industrial warehouse shelf. (Right) \tetra Picking an item from a stacked object scene in simulation.}
    \label{fig:teaser}
\end{figure}

Advances in robotic automation have led to improved grasping and object manipulation in industrial warehouses. However, common scenarios in unstructured environments still pose challenges to state-of-the-art systems. Single suction cups often struggle with occluded, irregularly shaped, or porous objects, leading to unstable or failed grasps. \autoref{fig:teaser} illustrates a typical scenario in industrial warehouses where a robot needs to pick an object in a stacked arrangement.

A typical pipeline during the Amazon Picking Challenge  (APC) consisted of a perception system that segments the scene and calculates grasp affordances based on visual images, followed by a robot that executes the pick without real-time adjustments \cite{Zeng2017MultiView}. However, object placement disturbances, perception errors, and complex scene configurations can reduce grasp success, particularly for robotic systems using vacuum grippers. During the picking process, the target object may shift, invalidating the initial affordance calculation, or obstructions may require repositioning other objects before grasping the target.

To address these challenges, we introduce \tetra, a novel multi-vacuum gripping strategy that improves upon traditional paradigms by mounting four suction cups on linear actuators and leveraging a previously unexplored sensing 
% \vspace{10mm}
modality for suction-based grasping: time-of-flight proximity sensors.

Proximity sensors are well-suited for reactive grasping applications, because they are resilient to occlusions, able to measure beyond immediate contact areas, and capable of real-time feedback. By equipping each linear actuator with a proximity sensor, our strategy enables the gripper to adapt to a large variety of object geometries in real time.

Further, proximity sensors can capture discriminative geometric features for local suction affordance estimation, augmenting grasp quality predictions beyond static affordance calculations. By providing continuous feedback during the grasping process, these sensors enable an actuated gripper to dynamically adjust in real-time, enhancing grasping success rates in cluttered and uncertain environments.

This approach outperforms single suction cup and fixed multi-suction cup strategies by securing complex geometries at multiple locations, enhancing grasping success rates. Additionally, real-time feedback enables the gripper to dynamically adjust to target object shifts during extraction, eliminating the need for robotic arm repositioning.

We make the following contributions:
\begin{itemize}
    \item A novel low-latency multi-suction actuated gripper design for adaptive bin picking, leveraging optical time-of-flight (ToF) proximity sensors and onboard machine learning capability.
    \item Develop a sensing-enabled suction cup for object manipulation and grasping, utilizing real-time ToF feedback.
    \item  A RL grasping policy for picking stacked objects based on multi-sensory input.
    \item Real-world experiments comparing our method to single suction grippers.
\end{itemize}

\begin{figure*}[t]
    \centering
    \begin{minipage}[b]{.6\textwidth}
        \centering
        \raisebox{0mm}{\includegraphics[trim=.0cm 2.0cm 5cm .0cm, clip,width=\textwidth]{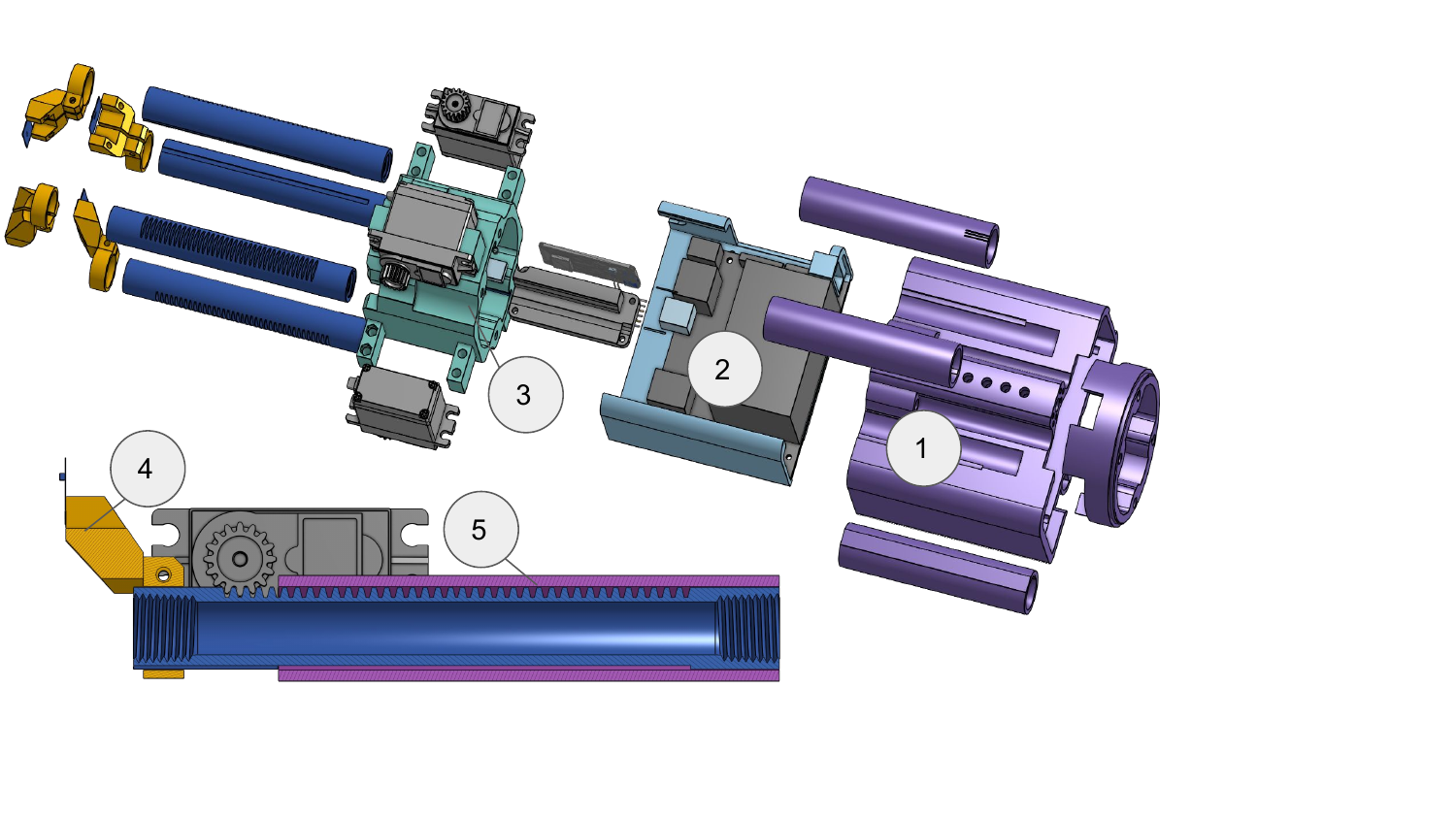}}
        \subcaption{}
        \label{fig:exp}
    \end{minipage}%
    \begin{minipage}[b]{.415\textwidth}
        \centering
        \includegraphics[trim=9.0cm 2.0cm 0.cm .0cm, clip,width=\textwidth]{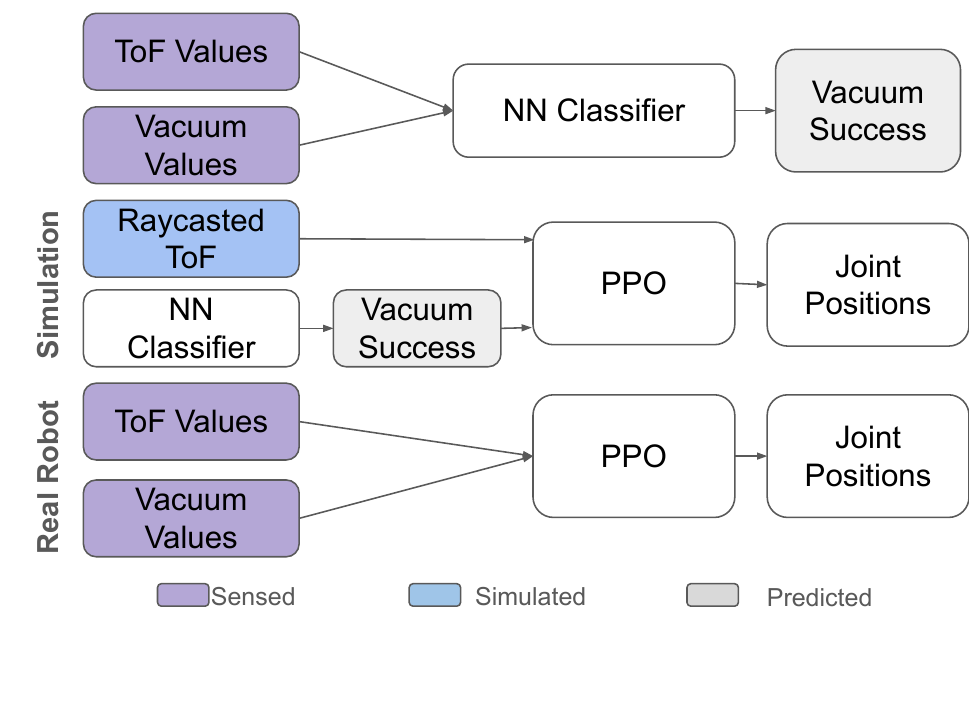}
        \subcaption{}
        \label{fig:system_architecture}
    \end{minipage}
     \caption{(a) Exploded view of the hardware design consisting of four suction cups which can be actuated individually. The housing contains a Jetson Nano Orin for onboard computing. (b) \tetra's sensor fusion and PPO-Based Control in Real and Simulated Environments}
\end{figure*}

\section{Related Work}

In proximity sensing applications, ToF sensors fall under the category of pretouch sensors - \SI{10}{\centi\meter} or less \cite{navarro.2011}. ToF sensors have been used for object manipulation because of their ability to accurately measure a wide range of materials without disturbing the target objects unlike tactile sensors that may cause the object to be misplaced. Lancaster et al. used the VL6180X \cite{Yang2017PreTouch}, \cite{Huang2018Visionless} to design a fingertip-sized sensor for the PR2's parallel jaw gripper and successfully demonstrated robust manipulation of a Rubik’s Cube. In our approach, we wish to leverage these techniques for stacked objects.

Suction grasping is commonly employed in warehouse fulfillment centers for pick-and-place tasks. The APC showcased the benefits of suction over parallel-jaw and multifinger grasping, as both winning teams (2015 and 2016) used low-profile grippers capable of reaching narrow spaces and lifting objects with a single point of contact \cite{Zeng2017MultiView}\cite{Hernandez2016TeamDelft}. The challenge for suction cups lies in achieving a secure seal on the target surface while generating enough force on the target object to deform the cup for effective adhesion. However, this force can disturb the target object, potentially moving it out of reach or invalidating prior affordance calculations. The linear actuators in \tetra allow for low-profile configurations as well as multiple points of contact where wrench resistance is required for large or heavy objects. The proximity sensors provide live adjustment during object manipulation.

In 2017 Hasegawa et al. \cite{Hasegawa2017ThreeFingered} tackled grasping in narrow, cluttered spaces using a three-fingered hand with integrated suction. The gripper combines finger-based grasping with suction, allowing it to adapt to different object shapes and sizes while improving grasp stability in constrained environments. Maggi et al. \cite{Maggi2022POLYPUS} presents the design and analysis of POLYPUS, an innovative robotic gripper that combines underactuation with vacuum-based grasping to enhance adaptability and handling capabilities. This gripper is engineered to manage objects with diverse shapes and surface characteristics, including both even and uneven geometries. The gripper introduces an innovative approach to object grasping, enhancing adaptability and stability in various environments. \tetra is inspired on these designs while providing low-profile solution such as the grippers that won the APC, which were specifically designed for accessing tight spaces.

Lee et al. \cite{Lee2024SmartSuction} developed an advanced robotic end-effector designed for both gripping and tactile exploration. This innovative suction cup features multiple internal chambers, each connected to a pressure sensor, enabling it to detect and adapt to various surface textures and contours. Borrowing from this idea, we connected \tetra to an industrial vacuum ejector capable of providing live feedback of the vacuum level at each suction cup.

Extracting objects from unstructured environments with a single robotic arm can lead to unintended object displacement or cause surrounding items to fall. A promising solution for such tasks is bimanual manipulation, as proposed by Grotz et al. \cite{grotz2025peract2}. However, this approach increases cost and complexity compared to our single-arm solution.

Recent advancements in deep learning has allowed researchers to develop algorithms for grasping synthesis using suctions cups. Dex-Net 3.0 introduced a compliant suction contact model that evaluates the quality of seals between suction cups and target surfaces, assessing the grasp's ability to withstand external forces \cite{Mahler2018DexNet3}. Utilizing this model, the authors generated a dataset of 2.8 million point clouds and corresponding grasp robustness labels, which was used to train a Grasp Quality Convolutional Neural Network (GQ-CNN) for identifying reliable suction grasp points in cluttered environments. Yang et al. improved upon this model by introducing object dynamics to the affordability calculation \cite{Yang2023DYNAMO}. 

The Transporter Network (Zeng et al.) is a vision-based model for robotic manipulation that infers spatial displacements by rearranging deep visual features, enabling efficient multi-step object rearrangement without explicit object representations. While it excels in structured tasks such as stacking, assembly, and deformable object manipulation, it lacks real-time feedback and adaptive control based on sensor inputs, making it less suitable for reactive grasping. 

Researchers have explored vacuum-based robotic grasping strategies for improving efficiency in cluttered environments through multi-suction cup mechanisms and learning-based grasp planning. Jiang et al. (2023) present a multiple-suction-cup vacuum gripper designed for grasping multiple objects simultaneously in cluttered scenes \cite{Jiang2023MultiSuction}. Their method leverages 3D convolutional affordance maps to determine optimal suction cup activation, significantly improving picking efficiency across different object categories. 

Complementing this work, Schillinger et al. (2023) propose a model-free grasping strategy for multi-suction cup grippers in robotic bin-picking applications \cite{Schillinger2023MultiSuction}. Their method employs a neural network to predict grasp quality and optimally configure suction cup activation without relying on explicit object models. This approach enhances the adaptability of suction-based grippers, making them more suitable for unstructured settings where object variability is high. In a different direction, Li and Cappelleri (2024) introduced Sim-Suction, a learning-based suction grasp policy trained on a large synthetic dataset of cluttered grasping scenarios \cite{Li2024SimSuction}. Their model, Sim-Suction-PointNet, predicts 6D suction grasp poses, leveraging point-wise affordance learning and zero-shot text-to-segmentation techniques. 

Together, these studies highlight the growing intersection of data-driven learning, affordance modeling, and multi-suction grippers, inspiring us to incorporate deep learning techniques in our approach to stacked object manipulation.

\section{\tetra System Architecture and Hardware Design}

The gripper is mounted on a Universal Robots UR16e, which is controlled by a desktop computer acting as the ROS master node. In the following sections we will describe the hardware behind \tetra.

\subsection{Compute}
Given that reactive grasping requires low latency, we selected the Jetson Orin Nano as the compute platform. It connects the gripper to the robot’s control desktop via Ethernet, handling sensor data collection, servo control, and deep learning inference locally. 

\subsection{Sensors}
The sensor used in our system is the VL53L5CX, an $8\times8$ multizone Time-of-Flight (ToF) ranging sensor with a wide field of view. It offers high accuracy, ranging capabilities down to \SI{2}{\centi\meter}, and a data rate of \SI{15}{\Hz}.
To maximize its effectiveness, we mounted the sensor at the end of the linear actuator rods via a 3D-printed housing, positioning it near the contact edge of the suction cup for precise proximity measurements. For robustness, the sensors are connected to a microcontroller, which communicates with the Jetson Orin Nano via USB.
These sensors play a critical role in our reactive grasping strategy, providing real-time depth feedback to dynamically adjust actuator positioning and ensure a secure seal.

\subsection{Vacuum System}

An industrial four-station vacuum ejector (Schmalz SCTSi unit with vacuum ejector type SCPS-15) provides suction for the gripper and is mounted on the robot structure. The unit converts compressed building air into a vacuum source, with a maximum vacuum of \SI{85}{\kilo\pascal}. Our measurements indicate that the vacuum sensor included with the unit operates at a data rate of approximately \SI{130}{\Hz}. The vacuum ejectors are controlled through ROS Simple Action Servers.

\subsection{Fabrication}
The gripper was constructed in a multi-step process to simplify fabrication. The body of the gripper was manufactured using ABS via 3D printing and bonded with ABS cement. As shown in \autoref{fig:exp}, the gripper body (Part 1) consists of six parts: the main body, the UR16e connector, and four linear actuator barrels. The Jetson Orin Nano (Part 2) is secured in a housing that is press-fitted into the gripper body.

Part 3 houses the servos, proximity sensor microcontroller, and servo controller, and is fastened to the gripper body with four bolts.

A cutout view of the linear actuator assembly is depicted in \autoref{fig:exp}. The assembly consists of four parts: the barrel, rod, servo, and proximity sensor housing (Part 4). The rod is an aluminum tubing extrusion with a machined rack and inlet/outlet ports (Part 5). In its compact form, the rod serves multiple functions: 1. 
It acts as a vacuum transport channel. 2. It functions as a structural rack. 3. It provides omnidirectional mechanical strength to resist bending moments due to the grasped object's weight. The linear actuators are driven by four high-torque servos, which are connected to the Jetson Orin Nano via a servo motor driver.

\section{Reactive Grasping Strategy}

Manipulating stacked objects requires a reactive grasping strategy, as objects may shift during the extraction process. Once the gripper is positioned in front of the target object, the control objective is to dynamically adjust the linear actuators to establish a secure seal on the suction cup(s), maximizing vacuum pressure to $ P_{\text{max}} = \SI{85}{\kilo\Pa}$ for a fully sealed condition. The system employs a reactive control strategy that integrates sensor feedback, adaptive actuator movements and vacuum ejector feedback to handle diverse grasping scenarios.

\subsection{Grasping Baseline}
\subsubsection{Single-Object Grasping}

A Proportional-Derivative (PD) controller adjusts the linear actuators based on Time-of-Flight (ToF) sensor feedback, ensuring that the fingers align with the object's surface for a stable suction grasp:
\begin{equation}
x_i(t) = K_p (d_i - d_{\text{target}}) + K_d \frac{d}{dt} (d_i - d_{\text{target}})
\end{equation}%
where $ d_i $ is the average of all current ToF sensor readings, \( d_{\text{target}} \) is the desired contact distance for suction engagement, and \( K_p, K_d \) are proportional and derivative gains.

\subsubsection{Stacked Object Handling}

\begin{figure}[H]
\centering
\includegraphics[trim=.0cm 5.0cm .0cm .0cm, clip, width=0.48\textwidth]{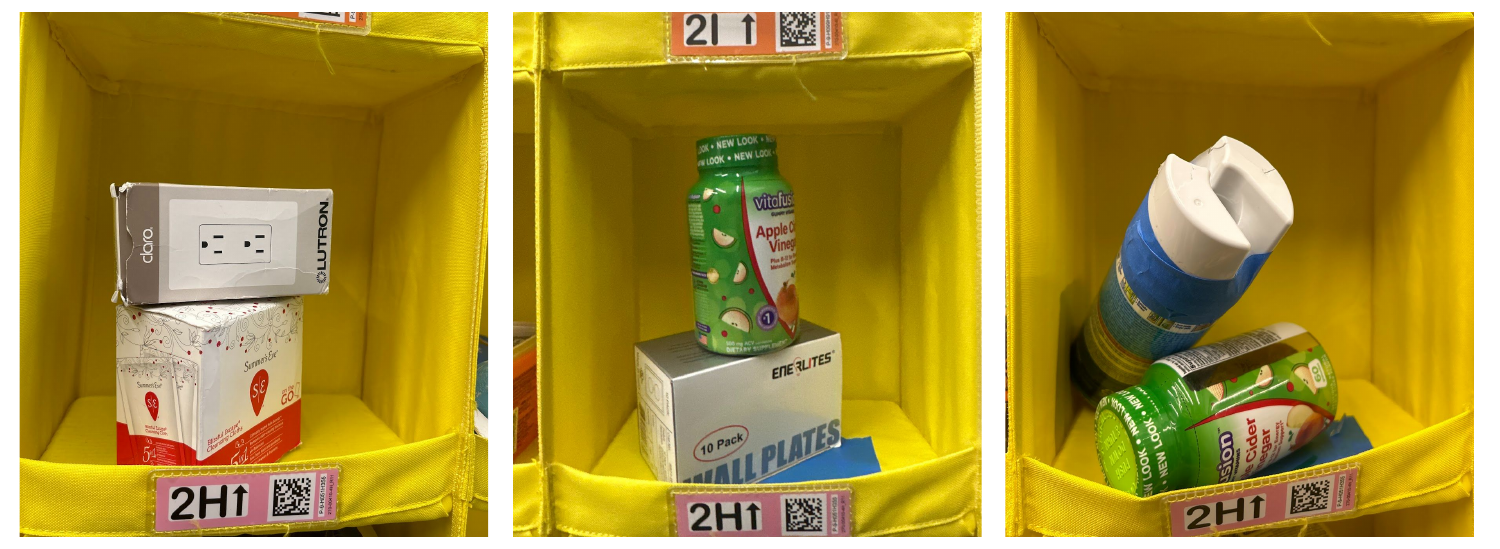}
\caption{Stacked object arrangement for real-world experiments}
\label{fig:stacked_objects_experiments}
\end{figure}

The objective of this experiment is to coordinate multiple linear actuators to manipulate a stacked scene (\autoref{fig:stacked_objects_experiments}). One actuator is responsible for grasping the target object, another actuator straightens it if necessary, and a third actuator pushes the stacked object away.

The angle of the target object is determined using proximity sensor readings from the grasping actuator. To assess whether the object needs straightening, the sensor data is divided into left and right halves, and the average depth values for each side are computed. If a significant imbalance is detected, the straightening actuator pushes the target object.

To demonstrate more complex conditions, a special case of the stacked object scenario experiment was conducted to demonstrate the advantage of multiple linear actuators. To increase the difficulty of sliding the stacked object, a retail item with a semi rigid hanging tab was used as the target object. This object can be seen inside the bin in \autoref{fig:hard_stacked}. 

\subsubsection{Blocked Object Handling}

In this experiment a blocking object does not allow direct access to the target object (\autoref{fig:blocked_object}). To ensure a successful grasp, the system stabilizes the target object by applying suction with one linear actuator before attempting to displace the blocking object. The displacement motion of the reciprocating linear actuator follows a sinusoidal function of increasing value. By first securing the target object and then using a controlled sinusoidal push, this strategy ensures that the target remains within reach while efficiently clearing obstructions.

\subsection{Grasping RL Policy}

Our RL policy follows \cite{andrychowicz2020learning}. 
We employed a RNN-LSTM Proximal Policy Optimization (PPO) policy to facilitate the stacked objects experiment. 
For high-quality data generation with realistic rendering, we implemented our method using Nvidia's Isaac Sim, while Nvidia's Warp was utilized as a high-performance GPU-accelerated raytracing framework \cite{Macklin2022Warp}. 
At each simulation timestep, the scene is converted into a mesh representation via Trimesh and PyTorch3D, subsequently processed by Warp for raytracing \cite{DawsonHaggerty2019Trimesh} \cite {Ravi2020PyTorch3D}.

To simulate successful suction, we collected approximately 8,000 samples of readings from proximity and vacuum sensors during real suction cup interactions while grasping a surface at angles between \SI{0}{\degree} and \SI{45}{\degree}. 
Using this dataset, we trained a neural network classifier that takes proximity sensor and vacuum readings as input and predicts whether suction is occurring \autoref{fig:system_architecture}.

The policy’s observation space is multi-modal and includes the joint positions of the servo motors, proximity sensor readings, and the binary outcome of successful suction based on vacuum levels. The action output is the desired servo positions, which are fed to a PD controller. A reward is provided when a finger successfully suctions onto the target object. Once the object is grasped, additional rewards are given for pushing the top object away and for bringing the target object closer to the gripper.
\section{Experimental Setup and Evaluation}

Our experimental setup features a cantilever-mounted Universal Robots UR16e, as shown in \autoref{fig:teaser}, positioned in front of a warehouse shelving unit containing multiple bins \cite{Grotz2023rss}. For evaluation, we selected a specific bin. During each trial, the object(s) are placed inside the bin, and the robot, controlled by a state machine, autonomously executes the pick request. During a pick attempt, the gripper ROS action server blocks the state machine while the pick operation is in progress, either successfully grasping the target object or timing out if the grasp attempt fails.

To showcase the versatility of the gripper, we conducted three sets of experiments: (1) simple pick, (2) stacked objects, and (3) blocked target object. The simple pick experiment evaluates the gripper’s ability to grasp a single object using sensor-based adjustments. The stacked objects experiment tests how the system handles multi-object scenarios, where the gripper must determine the correct grasp strategy based on object arrangement. The blocked target object experiment assesses the gripper’s ability to remove an obstructing object before attempting to grasp the intended target. All base case experiments use a single suction cup mounted on a centrally positioned linear actuator, aligned with the centroid of the target object.

\begin{figure}[H]
\centering
\includegraphics[width=0.5\textwidth]{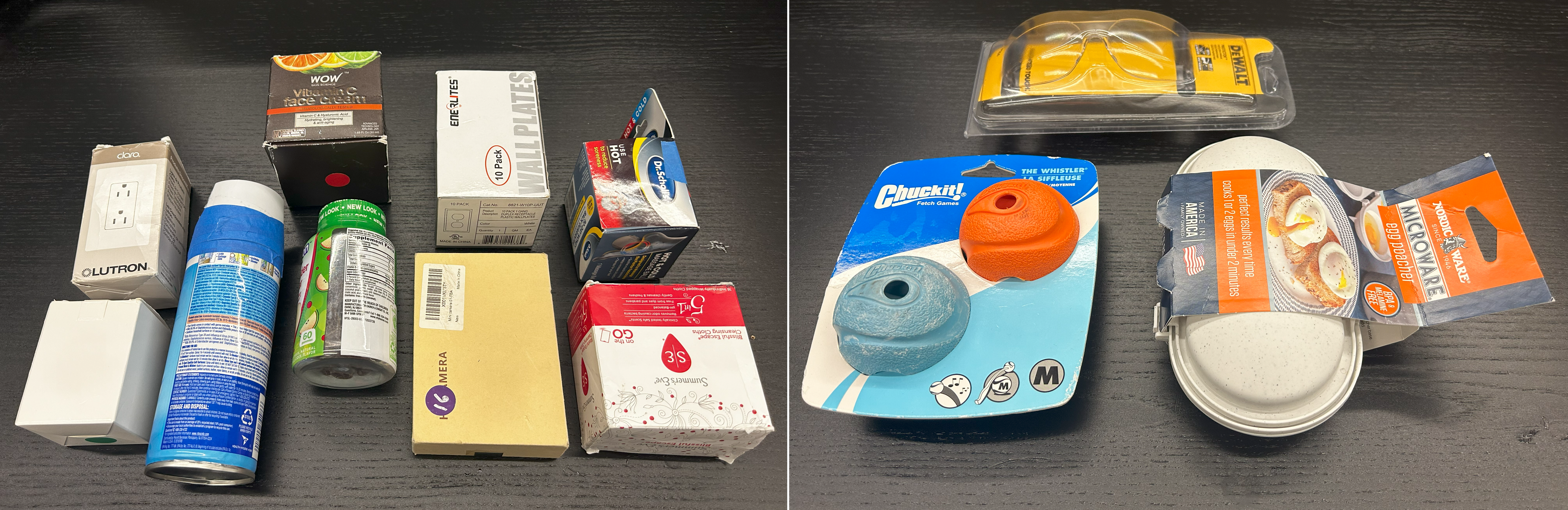}
\caption{Objects used in experiments}
\label{fig:objects}
\end{figure}

A total of 12 objects of varying sizes, shapes, and materials were used during the trials, as shown in \autoref{fig:objects}. The objects were selected to evaluate the gripper’s ability to handle different geometric and physical properties. This includes square and flat surfaces, which test the suction cup’s ability to form a complete seal, as well as cylindrical objects, which introduce curvature and potential challenges in achieving stable suction. Additionally, we included complex objects, such as highly irregular items, to assess the gripper’s robustness in handling difficult grasp scenarios. The selection ensures a comprehensive evaluation of the gripper’s adaptability across a wide range of real-world grasping challenges.
A grasp attempt is counted as successful if the object is securely attached to the gripper and successfully extracted out of the bin.

\subsection{Experiment I: Single-Object Grasping}
\begin{figure}[H]
    \centering
    \includegraphics[height=.32\textwidth] {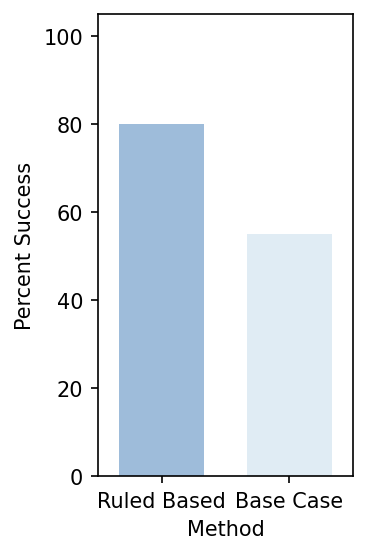}
    \caption{\tetra success rate for complex geometries.}
    \label{fig:succes_rate_single}
\end{figure}

Our first experiment involved grasping objects with complex geometries to assess \tetra’s capabilities. \autoref{fig:succes_rate_single} shows the results; in over 25 trials, \tetra achieved a \SI{80}{\percent} success rate.

\subsection{Experiment II: Stacked Object Handling}

\begin{figure}[H]
    \centering
    \includegraphics[height=.32\textwidth]{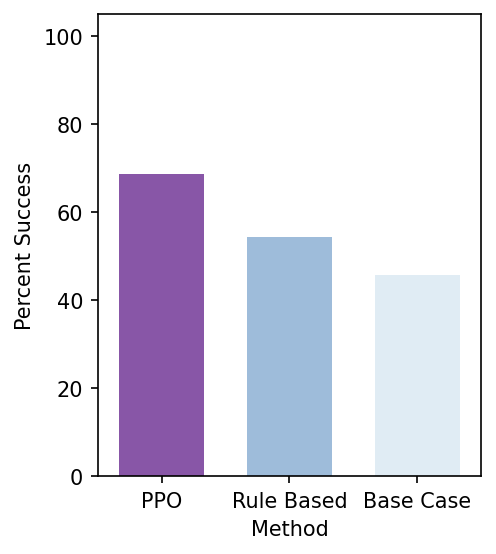}\hfill
    \caption{\tetra success rate comparison across different strategies for stacked objects.}
    \label{fig:success_rate_stacked}
\end{figure}

The results presented in \autoref{fig:success_rate_stacked} highlights the improvement in success rate with the use of multiple linear actuators and proximity sensors with respect to the base case which uses simply one linear actuator. Our PPO approach, achieving a success rate of \SI{68.57}{\percent}  across the experiments, outperforms all other methods by leveraging RL. The Rule-Based strategies, with a \SI{54.29}{\percent} percent success rate provide a moderate level of success but lacks the adaptability seen in learning-based approaches. 

This comparison suggests that PPO outperforms rule-based methods, reinforcing the advantage of learning-based control over predefined rules.

\begin{figure}[H]
    \centering
    \includegraphics[trim=.25cm .0cm .0cm .0cm, clip, height=.32\textwidth]{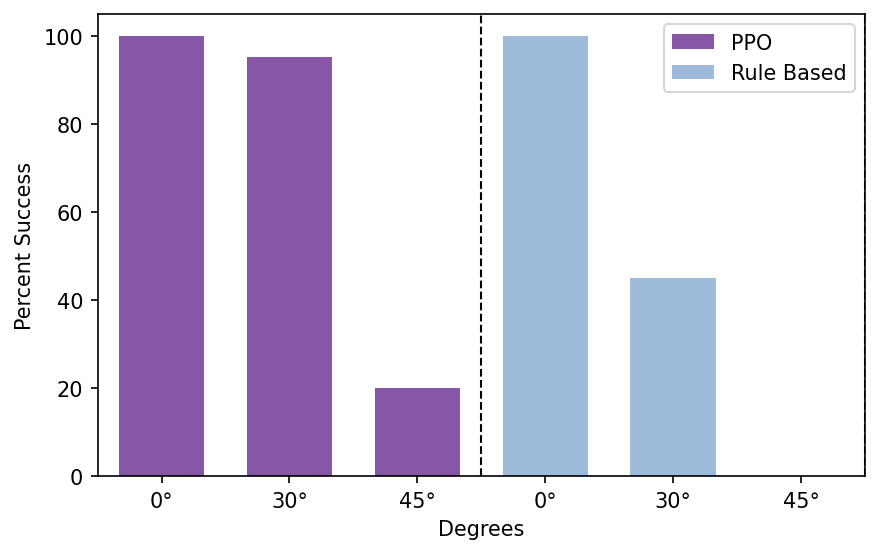}\hfill
    \caption{\tetra success rate comparison across different target object angles.}
    \label{fig:success_rate_stacked_angles}
\end{figure}

\autoref{fig:success_rate_stacked_angles} shows the success rate across different angles for PPO and Rule-Based control. Our RL policy presents an advantage over the rule based strategy as the angle of the target object sharpens.

\begin{figure}[H]
    \centering
    \includegraphics[trim=.0cm 8.0cm 12.0cm .0cm, clip, width=.45\textwidth]{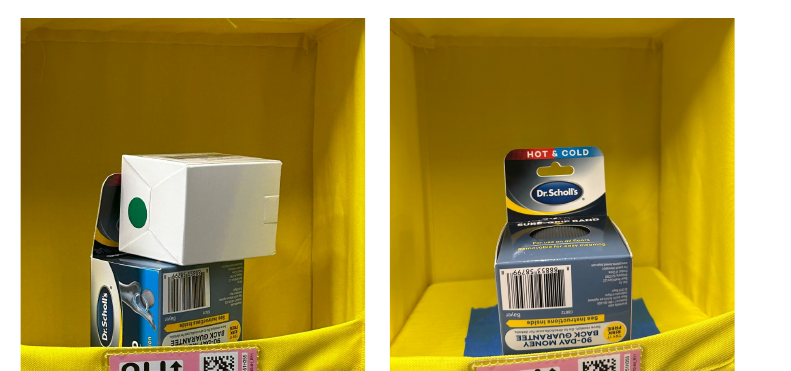}\hfill
    \caption{(Left) Before and after (Right) images of the complex grasping scenario for stacked objects.}
    \label{fig:hard_stacked}
\end{figure}

\begin{figure}[H]
\centering
\includegraphics[trim=.0cm 0cm .0cm .0cm, clip, width=0.32\textwidth]{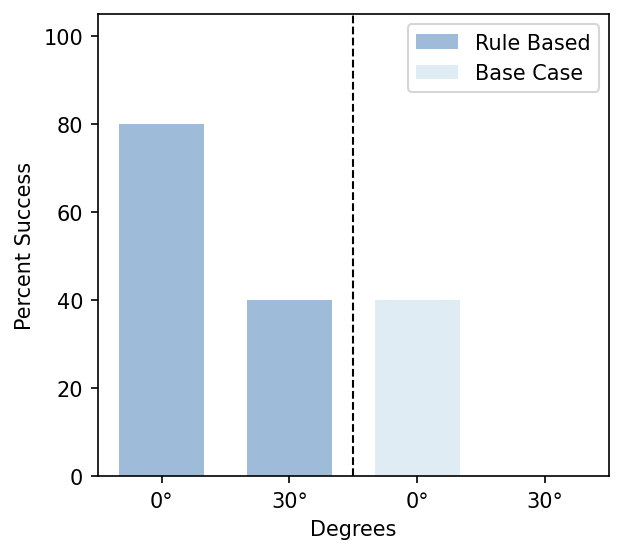}
\caption{Complex grasping scenario for stacked objects experiment results.}
\label{fig:hard_stacked_sucess}
\end{figure}

To demonstrate the advantages of multiple linear actuators, we performed an additional experiment (\autoref{fig:hard_stacked}), where the bottom object has an extended back, increasing the likelihood of the top object remaining in place. Success rates are presented in \autoref{fig:hard_stacked_sucess}. 

\subsection{Blocked Object Handling}

\begin{figure}[H]
    \centering
    \begin{minipage}[b]{0.2\textwidth}
        \centering
        \raisebox{5mm}{\includegraphics[trim=.0cm .0cm 1.cm .0cm, clip,width=\textwidth]{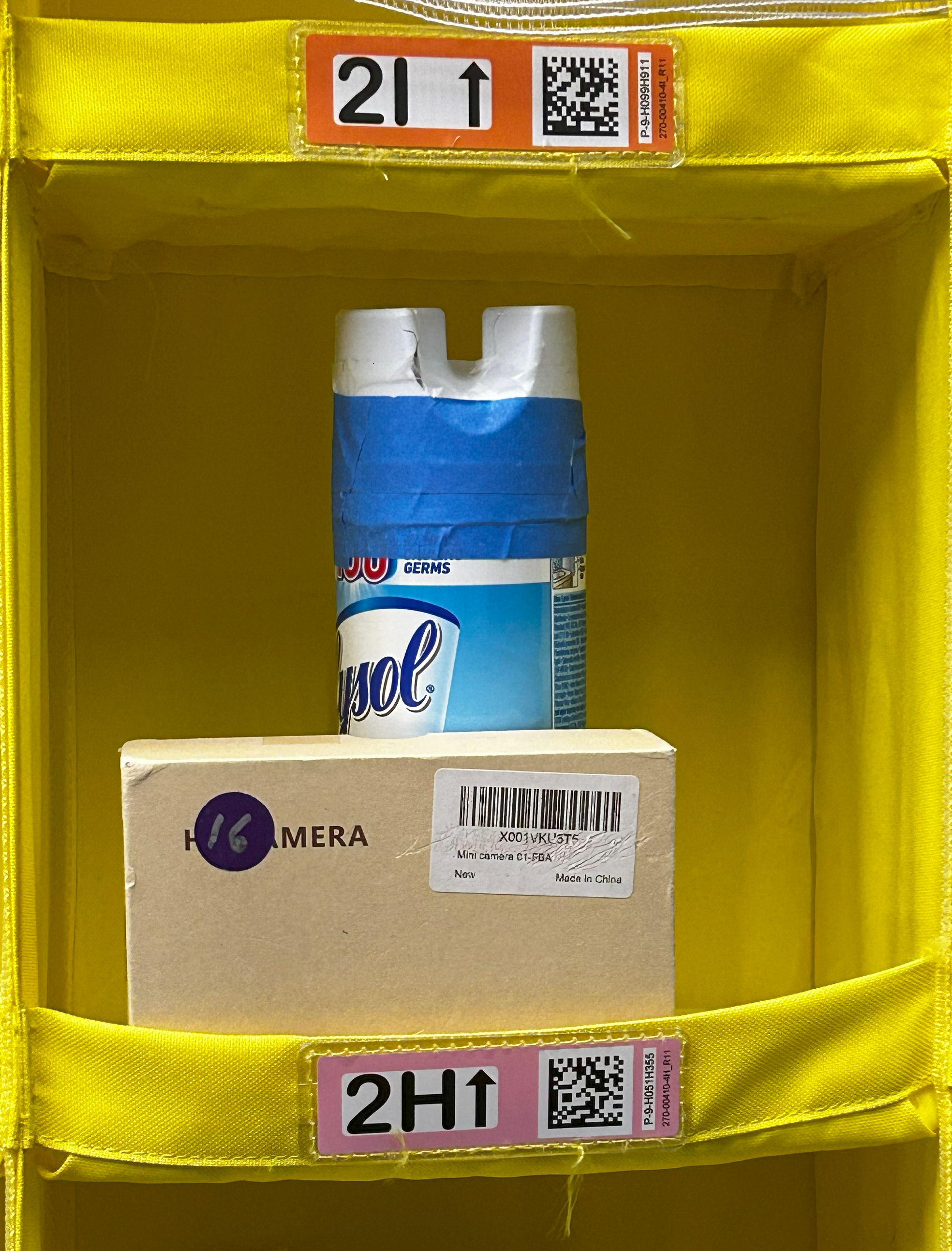}}
        \subcaption{}
        \label{fig:blocked_object}
    \end{minipage}%
    % \hfill
    \begin{minipage}[b]{0.2\textwidth}
        \centering
        \includegraphics[width=\textwidth]{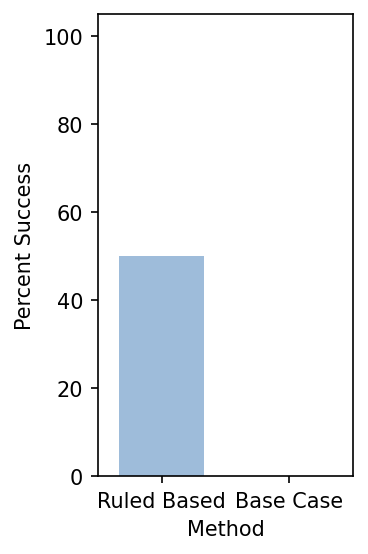}
        \subcaption{}
        \label{fig:blocked_object_success}
    \end{minipage}
     \caption{(a) Blocked object configuration. (b) Success rate of blocked object experiment.}
\end{figure}

The third experiment demonstrates the versatility of the gripper in complex grasping scenarios where a blocking object prevents direct access to the target (\autoref{fig:blocked_object}). Using a rule-based strategy, the system successfully retrieved the target object in 50 percent of trials across a 20-trial experiment. This result highlights the effectiveness of the approach in stabilizing the target object before executing a controlled displacement of the obstruction.

\section{Discussion}

The results of our experiments demonstrate the advantages of multi-actuated suction grippers in complex grasping scenarios, particularly in stacked and obstructed object configurations. Traditional single-suction grippers rely on precomputed affordance maps and static grasp planning, which can fail when objects shift or when occlusions prevent direct suction. In contrast, \tetra's combination of linear actuators and time-of-flight (ToF) sensors enables real-time adjustments, allowing for reactive grasping strategies that adapt to environmental uncertainties.

One key finding is that RL-based control significantly improves grasping performance over rule-based approaches when the target surface is not normal to the suction cup. By leveraging sensor feedback, an RL policy is able to more smoothly adjust actuator positions, optimizing contact forces and suction effectiveness. In the stacked object scenario with a tilt of 30 degrees for the target object, the RL-based method outperformed rule based by 50 \%.
However, a key concern is the generalization limitations of PPO. Since the policy was trained on a specific scenario, there is uncertainty regarding its ability to adapt to unseen object poses, varying surface properties, or broader warehouse conditions. Future work should explore further domain randomization and alternative RL architectures to enhance generalization and ensure robust performance across diverse grasping tasks.

We were able to show the advantage of using multiple linear actuators in complex conditions such as picking an occluded object or retrieving an object obstructed by another. However, we did not train RL policies specifically for these tasks. It is feasible to train a policy that generalizes across multiple scenarios, potentially improving adaptability in complex environments. Future work could explore multi-task learning or curriculum learning strategies to develop policies capable of handling a broader range of grasping challenges.

Several limitations remain, the gripper is stationary during the picking interaction in order to isolate and assess its performance independently. However, incorporating gripper motion during interaction would likely enhance grasping success. Although more complex, an RL policy that jointly optimizes both gripper and arm positioning would likely outperform the results presented in this paper.

Furthermore, further tests on an object covered with a plastic bag, using industrial suction cups for bags, did not result in successful grasps.

\section{Conclusions}

We introduced \tetra, a novel multi-actuated suction gripper strategy designed for reactive grasping in unstructured environments. By integrating four linear actuators with ToF sensors, \tetra enables adaptive grasping strategies that dynamically respond to object placement and occlusions. We evaluated its performance in stacked and obstructed object scenarios, demonstrating that reinforcement learning-based control strategies improve grasping success by 22.86\% over a traditional single-suction gripper. Additionally, \tetra proved effective in complex conditions, successfully grasping objects in cases where a single-suction gripper was physically unable to perform the task.

Our findings highlight the advantages of multi-actuated, sensor-driven suction grasping, particularly in warehouse-style environments where clutter and varying object geometries present significant challenges. Future work will explore further optimization of control policies and the integration of additional sensing modalities to enhance robustness.

\section*{ACKNOWLEDGMENT}

This work has been supported by Amazon under the research project "Robotic Manipulation in Densely Packed Containers". We would also like to thank our colleague Entong Su for her assistance in setting up the simulation environment and Kory Dean from the University of Washington College of Engineering for his machining expertise.

\bibliographystyle{unsrt}
\bibliography{references}


%%%%%%%%%%%%%%%%%%%%%%%%%%%%%%%%%%%%%%%%%%%%%%%%%%%%%%%%%%%%%%%%%%%%%%%%%%%%%%%%

%%%%%%%%%%%%%%%%%%%%%%%%%%%%%%%%%%%%%%%%%%%%%%%%%%%%%%%%%%%%%%%%%%%%%%%%%%%%%%%%

%%%%%%%%%%%%%%%%%%%%%%%%%%%%%%%%%%%%%%%%%%%%%%%%%%%%%%%%%%%%%%%%%%%%%%%%%%%%%%%%
%\section*{APPENDIX}

%\section*{ACKNOWLEDGMENT}

\end{document}